# A new framework for experimental design using Bayesian Evidential Learning: the case of wellhead protection area

Robin Thibaut[a]*, Eric Laloy[b], Thomas Hermans[a]

*[a]Ghent University, Belgium (*Corresponding author: Robin.Thibaut@UGent.be, Thomas.Hermans@UGent.be)*

*[b]SCK CEN, Belgium (Eric.Laloy@sckcen.be)*

**Abstract:** Decisions related to groundwater management such as sustainable extraction of drinking water or protection against contamination can have great socio-economic impacts. Ideally, a complete uncertainty analysis should be performed to foresee all possible outcomes and assess any risk. Uncertainties arise from our limited understanding of the involved physical processes and the scarcity of measurement data, whether directly or indirectly related to the physical parameters of interest. In this contribution, we predict the wellhead protection area (WHPA, target), the shape and extent of which is influenced by the distribution of hydraulic conductivity (K), from a small number of tracing experiments (predictor). Our first objective is to make stochastic predictions of the WHPA within the Bayesian Evidential Learning (BEL) framework, which aims to find a direct relationship between predictor and target using machine learning. This relationship is learned from a small set of training models (400) sampled from the prior distribution of K. The associated 400 pairs of simulated predictors and targets are obtained through forward modelling. Newly collected field data can then be directly used to predict the approximate posterior distribution of the corresponding WHPA, avoiding the classical step of data inversion. The uncertainty range of the posterior WHPA distribution is affected by the number and position of data sources (injection wells). Our second objective is to extend BEL to identify the optimal design of data source locations that minimizes the posterior uncertainty of the WHPA. This can be done explicitly, without averaging or approximating because once trained, the BEL model allows the computation of the posterior uncertainty corresponding to any new input data. We use the Modified Hausdorff Distance (MHD) and the Structural Similarity (SSIM) index metrics to estimate the posterior uncertainty range of the WHPA. As expected, increasing the number of injection wells effectively reduces the derived posterior WHPA uncertainty, because the breakthrough curves store information on a large area of the K field surrounding the pumping well. Our approach can also estimate which injection wells are more informative than others, as validated through a k-fold cross-validation procedure. Overall, the application of BEL to experimental design makes it possible to identify the data sources maximizing the information content of any measurement data within limited budget constraints, and at limited computational costs.

## 1. Introduction

Wellhead Protection Area (WHPA) is defined as the zone surrounding a pumping well where human activities are restricted to protect the water resources (Goldscheider, 2010), generally based on the amount of time harmful contaminants within the area will take to reach the pumping well (according to local regulation). It depends on the flow velocity in the subsurface around the well, and it can be computed numerically by particle tracking or transport simulation, or in practice by tracer tests (Goldscheider, 2010; Dassargues, 2020). Typically, a groundwater model is first calibrated against field data and the WHPA is computed from the calibrated model. The delineation of such zones can have great socioeconomic impact in populated areas where land occupation is of prime concern. As such, the outcome of each possible event should be quantified to best support decision making.

Traditional deterministic calibration methods to compute the dimensions of this area might not be appropriate (Zhou et al., 2014; Kikuchi et al., 2015), as they do not consider the uncertainty of such prediction problems, arising mostly from our limited knowledge about the heterogeneity of the subsurface. Instead, stochastic methods should be used to assess the full range of possible outcomes, allowing a thorough risk analysis as the basis for decision making (de Barros et al., 2012; Zhou et al., 2014; Linde et al., 2017). The drawback of stochastic methods is their computational cost. Generally, the solution is obtained by iterative methods requiring many runs of the forward problem such as Markov chain Monte Carlo (McMC) methods (Laloy and Vrugt, 2012, Vrugt, 2016) or stochastic optimization (Hu et al., 2001; Hermans et al., 2015) to fit the observed data. The number of iterations increases with the complexity of the mathematical models describing the phenomena at hand and the number of parameters describing the subsurface models, discouraging uncertainty analysis or sensitivity analyses of groundwater models in practical applications.

Due to their design, these stochastic methods do not only allow to compute a forecast range, but they can also lead to experimental (or optimal) design for optimization of operations within limited budget constraints. Experimental design is generally defined as finding the data set that minimizes the uncertainty of a specific prediction, which can be expressed as maximizing or minimizing a data utility function (e.g., Kikuchi et al., 2015). However, the computational burden is even more important as experimental design supposes that the observed data is not known yet, requiring in theory to find the stochastic solution to the inverse problem for any possible outcome of the unknown data (e.g., Leube et al., 2012; Neuman et al., 2012). To make practical applications tractable, two main simplifications have been proposed: (1) Bayesian Model Averaging (BMA), (Raftery et al., 2005; Vrugt and Robinson, 2007; Tsai and Li, 2008; Wöhling and Vrugt, 2008; Kikuchi et al., 2015; Pham and Tsai, 2016; Samadi et al, 2020), and (2) surrogate modelling (Razavi et al., 2012; Laloy et al, 2013; Asher et al., 2015; Babaei et al., 2015; Zhang et al., 2020).

The BMA approach is extensively described in Kikuchi et al. (2015), Raftery et al. (2005) or Samadi et al. (2020). In essence, BMA is a preposterior estimation technique allowing to estimate the expected value of the chosen data utility function by computing the average of several ensemble-generated realizations of the prospective data set, instead of calculating the full posterior. An example is presented in Kikuchi et al. (2015) who further extended the approach to propose a novel experimental design approach in hydrology, called Discrimination-Inference (DI), that can be used for conceptual and predictive discrimination. The latter is based on the effect of additional data collection on prior and posterior probability distributions, measured by the Kullback-Leibler divergence. Kikuchi and co-workers first calibrated $n$ sets of model parameters conditioned on existing data to populate an input matrix through McMC sampling. Having obtained a set of randomly sampled parameters, they applied forward modelling to produce data realizations to estimate the data utility function using BMA. In their approach, the posterior distribution of the prediction, which is at the heart of the data utility function, is never computed.

Another approach to reduce the computational cost of experimental design is to use surrogate models. Razavi et al. (2012) proposed an extensive review of surrogate modelling in the water resources field. The surrogate model of a computationally intensive process is used to estimate either an objective function, constraints, or both. Razavi et al. (2012) state that surrogate modelling can be divided in two main categories: high-fidelity response surface models and low-fidelity models, fidelity referring to the level of realism of simulation models. High-fidelity response surface models emulate the original model's numerical outputs. This type of methods consists of either statistical models or empirical data-driven models, discerning a relationship between model parameters and one or several model response variables, using techniques such as, e.g., kriging (Báu and Mayer, 2006; Garcet et al., 2006), artificial neural networks (Yan and Minsker, 2006; Kourakos and Mantoglou, 2009), radial basis functions (Regis and Shoemaker, 2005) or polynomial chaos expansion (Laloy et al., 2013) to name a few. Low-fidelity models are physically based. They are essentially simplified, less faithful versions of their computationally demanding parent models. In order to apply lower-fidelity models in practice, the response of the lower-fidelity model needs to be "reasonably close" to the response of the original model. Razavi et al. (2012) note that surrogate modelling becomes less beneficial or even impractical as the number of model variables increases, and consistently leads to a reduction in accuracy of the analyses. Once an appropriate surrogate model is available, it can be used to efficiently solve stochastic inversion and experimental design problems at a limited computational cost. However, the approximation can lead to significant bias in the prediction and therefore to erroneous estimation of the data utility function (Razavi et al., 2012; Laloy et al, 2013; Asher et al., 2015; Babaei et al., 2015; Zhang et al., 2020).

In this work, we propose an alternative for solving experimental design studies using the Bayesian Evidential Learning (BEL) framework (Hermans et al., 2016, 2018; Scheidt et al., 2018) within the context of WHPA estimation from tracing experiment breakthrough curves. BEL, as any posterior prediction approach, is based on the definition of a realistic prior distribution of the model parameters. Through forward modelling, a limited set of associated WHPA's (target, $\mathbf{h}$) and breakthrough curves (predictor, $\mathbf{d}$) are generated. The key idea of BEL is to find a direct relationship between $\mathbf{d}$ and $\mathbf{h}$ in a reduced dimensional space with machine learning. Then, given a new measured predictor $\mathbf{d}_*$, this relationship is used to infer the posterior probability distribution of the target, without the need for a computationally expensive inversion. The posterior distribution of the target is then sampled and backtransformed from the reduced dimensional space to the original space to predict posterior realizations of $\mathbf{h}$ (WHPA) given $\mathbf{d}_*$.

Previous works have shown the ability of BEL to predict the posterior distribution of predictions in many contexts including geothermal systems (Hermans et al., 2018; Athens and Caers; 2019), contaminant transport (Scheidt et al., 2015; Satija and Caers, 2015) and geophysical inversion (Hermans et al., 2016; Michel et al., 2020a, 2020b). It has been validated against rejection sampling (Scheidt et al., 2015), McMC algorithms (Michel et al., 2020a, 2020b) and field data (Hermans et al., 2019). In this work, we show that BEL also offers an efficient alternative for solving optimal design problems.

The main advantage of BEL in this context is that the inferred relationship can be used for any data set consistent with the prior distribution. We thus extend the capabilities of the BEL framework to solve the optimization problem of finding the predictor $\mathbf{d}_*$ which reduces the uncertainty on the prediction of the target, by defining a suitable data-utility function to minimize. This allows us to determine the optimal location of any injection well, with logistic and economic issues at stake. In contrast to BMA, we compute the full posterior at a reduced cost, and the approach does not use any simplifications of the forward model as with surrogate modelling.

The objective of this work is two-fold: firstly, we demonstrate the prediction capabilities of BEL for the WHPA estimation. Secondly, we introduce an experimental design approach based on BEL to the identification of informative hydrologic data. The remainder of this paper is organized as follows.

Section 2 develops the theoretical basis of our framework before Section 3 demonstrates the capabilities of BEL for prediction and uncertainty quantification. Section 4 then illustrates how we design the optimization under the BEL framework to make the best choice in the injection well location. Finally, Section 5 discusses the limitations of the approach and Section 6 provides a short conclusion.

## 2. Methodology
### 2.1. Bayesian Evidential Learning

The goal of BEL is to infer the posterior probability distribution $p(\mathbf{h}|\mathbf{d}_{obs})$ of a target $\mathbf{h} \in \mathbb{R}^t$, conditioned by an observed predictor $\mathbf{d}_{obs} \in \mathbb{R}^p$, by training a statistical model given a series of examples of both $\mathbf{d}$ and $\mathbf{h}$. Target and predictor are real, multi-dimensional random variables. For real applications in geosciences, we generally do not have multiple realizations of $\mathbf{d}$ or $\mathbf{h}$. However, we can rely on realistic distributions of the subsurface model parameters $\mathbf{M}$, to generate the prior distribution of $\mathbf{d}$ and $\mathbf{h}$ from which the statistical relationship can be learned (Hermans et al., 2016, 2018; Scheidt et al., 2018). Datasets encountered in geosciences can be high-dimensional, sparse, plagued by noise and possibly by multicollinearity. The key process of BEL is to work in a reduced-dimension space alleviating these issues. In this work, the dimensionality reduction process is two-fold. First, Principal Component Analysis (PCA) is applied to both target and predictor to aggregate the correlated variables into a few independent Principal Components (PC's) (Meloun and Militký, 2012). Next, Canonical Correlation Analysis (CCA) reduces the dimensionality of the PC's and transforms the two sets into pairs of Canonical Variates (CV's) independent of each another. While PCA seeks to maximize reconstruction of original variables, CCA finds directions for two variables to maximize their correlation. A CV in one set has the highest correlation with its corresponding CV in the other set. These solutions depend both on correlations among variables in each set and on correlations between predictor and target sets (Meloun and Militký, 2012). Hence, CCA corresponds to the learning step of the procedure. The interest of CCA is that the resulting bivariate distributions of pairs can be characterized. The most commonly used multivariate joint probability density function for continuous variables is the Multivariate Gaussian (MG) distribution (Murphy, 2012). One reason is that it can be fully defined by its first two moments, the mean $\boldsymbol{\mu}$ and covariance $\boldsymbol{\Sigma}$, that can be directly estimated from data. Note that it is not a requirement for BEL as other methods such as Kernel Density Estimation (KDE) can be used (e.g., Scheidt et al., 2018; Hermans et al., 2019; Michel et al., 2020a). Nevertheless, using the MG distribution simplifies the problem and is thus favored in this work. However, it requires to verify some assumptions. Indeed, the posterior distribution of $\mathbf{h}$ conditioned on $\mathbf{d}$ is Gaussian if the two sets of variables are jointly Gaussian. In particular, if the linear correlation between $(\mathbf{h}^c, \mathbf{d}^c)$ pairs (the superscript $c$ denotes the canonical space) with $q$ CV each is sufficiently strong, then the following can be done. Let $\mathbf{G} \in \mathbb{R}^{q \times q}$ be the Ordinary Least Square (OLS) solution mapping $\mathbf{h}^c$ to $\mathbf{d}^c$, such that $\mathbf{G}\mathbf{h}^c = \mathbf{d}^c + \boldsymbol{\epsilon}$. If the CV's of the predictor and the target sets are both normally distributed, then analytic MG inference can be performed to infer directly $p(\mathbf{h}^c|\mathbf{d}^c_{obs})$.

Suppose $p(\mathbf{h}^c|\mathbf{d}^c_{obs})$ is jointly Gaussian with covariance matrix (Murphy, 2012)

$$\boldsymbol{\Sigma} = \begin{pmatrix} \boldsymbol{\Sigma}_{\mathbf{h}^c\mathbf{h}^c} & \boldsymbol{\Sigma}_{\mathbf{h}^c\mathbf{h}^c}\mathbf{G}^T \\ \mathbf{G}\boldsymbol{\Sigma}_{\mathbf{h}^c\mathbf{h}^c} & \mathbf{G}\boldsymbol{\Sigma}_{\mathbf{h}^c\mathbf{h}^c}\mathbf{G}^T + \boldsymbol{\Sigma}^*_{\mathbf{d}^c} \end{pmatrix} = \begin{pmatrix} \boldsymbol{\Sigma}_{\mathbf{h}^c\mathbf{h}^c} & \boldsymbol{\Sigma}_{\mathbf{h}^c\mathbf{d}^c} \\ \boldsymbol{\Sigma}_{\mathbf{d}^c\mathbf{h}^c} & \boldsymbol{\Sigma}_{\mathbf{d}^c\mathbf{d}^c} \end{pmatrix}. \qquad 1$$

The precision matrix $\boldsymbol{\Lambda} \stackrel{\text{def}}{=} \boldsymbol{\Sigma}^{-1}$ is

$$\boldsymbol{\Lambda} = \begin{pmatrix} \mathbf{G}^T\boldsymbol{\Sigma}_{\mathbf{h}^c\mathbf{h}^c}\mathbf{G} + \boldsymbol{\Sigma}^{*-1}_{\mathbf{d}^c} & -\mathbf{G}^T\boldsymbol{\Sigma}^{-1}_{\mathbf{h}^c\mathbf{h}^c} \\ -\boldsymbol{\Sigma}^{-1}_{\mathbf{h}^c\mathbf{h}^c}\mathbf{G} & -\boldsymbol{\Sigma}^{-1}_{\mathbf{h}^c\mathbf{h}^c} \end{pmatrix} = \begin{pmatrix} \boldsymbol{\Lambda}_{11} & \boldsymbol{\Lambda}_{12} \\ \boldsymbol{\Lambda}_{21} & \boldsymbol{\Lambda}_{22} \end{pmatrix}. \qquad 2$$

The posterior conditional $p(\mathbf{h}^c|\mathbf{d}^c_{obs}) = \mathcal{N}\left(\mathbf{h}^c \middle| \boldsymbol{\mu}_{\mathbf{h}^c|\mathbf{d}^c_{obs}}, \boldsymbol{\Sigma}_{\mathbf{h}^c|\mathbf{d}^c_{obs}}\right)$ has parameters

$$\Sigma_{\mathbf{h}^c|\mathbf{d}^c_{obs}} = \Lambda_{11}^{-1}, \qquad 3$$

$$\mu_{\mathbf{h}^c|\mathbf{d}^c_{obs}} = \Sigma_{\mathbf{h}^c|\mathbf{d}^c_{obs}} \left( \Lambda_{11}\mu_{\mathbf{h}^c} - \Lambda_{12}\left(\mathbf{d}^c_{obs} - \mu^*_{\mathbf{d}^c}\right) \right). \qquad 4$$

Note that the posterior covariance (Eq. 3) does not depend on the observed value $\mathbf{d}^c_{obs}$, and that computing the posterior mean (Eq. 4) is simply a linear operation, given the precomputed posterior covariance. Each term in the above equations can be estimated with the training data in canonical space, under the normality and linear correlation assumptions. The $\Sigma^*_{\mathbf{d}^c}$ and $\mu^*_{\mathbf{d}^c}$ variables are respectively a covariance term and a term representing deviations from the mean, both arising from the imperfect OLS fitting and the noise present in the predictor set. Note that histogram transformation can be used to ensure the normality of the distributions (Satija and Caers, 2015). We refer to Scheidt et al. (2018) for information about how to approximate $\Sigma^*_{\mathbf{d}^c}$ and $\mu^*_{\mathbf{d}^c}$.

Once the first two moments of the MG $p(\mathbf{h}^c|\mathbf{d}^c_{obs})$ are known, sampling from it is straightforward. As the drawn samples belong to the canonical space and are thus not interpretable as such, they are backtransformed to the original physical space $\mathbb{R}^t$.

If the linear correlation and normality assumptions are too strongly violated, KDE can be considered to approximate $p(\mathbf{h}^c|\mathbf{d}^c_{obs})$ (e.g., Hermans et al., 2019). However, it requires to choose the kernel and define the corresponding bandwidth, which can be objectively estimated using the number of training samples in the vicinity of the observed data (e.g., Michel et al., 2020a). Note also that the BEL framework can handle noise propagation by estimating the data covariance matrix using Monte Carlo simulation (Hermans et al., 2016).

### 2.2. WHPA prediction

In this section we propose a method to predict with BEL the posterior distribution of an unknown WHPA given observed BC's. Distinct tracers originate from a maximum number $s$ of data sources (injection wells) at different locations around the pumping well. Both target and predictor are high-dimensional, and their relation is non-linear. In order to learn a direct relationship, both variables are first pre-processed, and their dimensionality reduced, before training the model and performing the multivariate regression.

#### *2.2.1. Pre-processing*
#### 2.2.1.1. Predictor

The predictor is a set of BC's measured at the pumping well, resulting from $n$ solute transport simulations. Each BC is a one-dimensional time series of concentration, with one concentration value for each time-step of each simulation. Since the used groundwater solute transport software (MT3D-USGS, Tonkin et al., 2016) has an adaptive time-step, the $n$ models do not produce BC's of the same dimension. Therefore, we interpolate the breakthrough curves and extract the corresponding concentration at $k$ pre-defined time steps, and each simulation results in a data set of dimensions $(\lambda \times k)$. $\lambda$ is the number of data sources used ($1 \leq \lambda \leq s$). For $n$ simulations, the predictor matrix is $\mathbf{B}^\lambda$ ($n \times \lambda \times k$). The last dimension of $\mathbf{B}^\lambda$, with $k$ elements, only contains concentration values ($\frac{kg}{m^3}$). As long as they share the same time steps, the time dimension becomes irrelevant. The information that needs to be stored is the shape of the curves and the magnitude of the concentration values. For each sample in $\mathbf{B}^\lambda$, individual curves of each well are concatenated. The shape of $\mathbf{B}^\lambda$ becomes $(n \times \lambda \cdot k)$, where each row contains the raw predictor in physical space for each simulation. PCA can then be performed to obtain the principal components (PC's). Let $\beta = \min(n, \lambda \cdot k)$, i.e., the maximum possible number of PC's stored in the $(n \times \beta)$-matrix $\mathbf{D}_\beta$. The last pre-processing step for the predictor is to choose an adequate PC number $1 \leq \delta < \beta$ to reduce the dimension of the predictor while preserving enough variance to adequately represent the original dataset. The final predictor training set is the $(n \times \delta)$-matrix

$$\mathbf{D}_\delta = \{\mathbf{d}_{i,j} \mid 1 \leq i \leq n;\ 1 \leq j \leq \delta\}. \qquad 5$$

PCA is only performed on the training set. The PCA coefficients are kept in order to project the actual observation set to the PC space later on during prediction, assuming that the test set is consistent with the training set. The verification of this assumption falls out of the scope of this contribution. For field cases, however, the experimenter must check for inconsistency. See Scheidt et al. (2018) and Hermans et al. (2018) for more details.

### 2.2.1.2. Target

The targets of this study are wellhead protection areas, determined based on the 30-days endpoints of $b$ backtracked particles ending up in the pumping well. For $n$ forward model simulations, the raw output is the $(n \times b \times 2)$-matrix

$$\mathbf{E} = \{\mathbf{e}_1, \mathbf{e}_2 \ldots \mathbf{e}_{i-1}, \mathbf{e}_i \mid 1 \leq i \leq n\}, \qquad 6$$

$$\mathbf{e}_i = \{\{\mathrm{x},\mathrm{y}\}_j \mid 1 \leq j \leq b\}, \qquad 7$$

with $(\mathrm{x},\mathrm{y})$ the coordinate of a particle endpoint. The latter is actually 3-dimensional, but since a single layer is considered in this work, the depth dimension is discarded. As the number of particles $b$ in the discretized space is increased, so is the resolution of the 2-dimensional WHPA delineation. The pattern of a WHPA is controlled by the heterogeneities of the hydraulic conductivity field in the vicinity of the pumping well. It is defined by the area of the polygon whose vertices are a particular ordering of the particle's endpoints. The output file of the used particle tracking software (MODPATH 7, Pollock (2016)) does not provide such a list of sorted coordinates. We therefore compute the WHPA polygon by solving the traveling salesman problem (TSP), a combinatorial optimization problem (COP) which aims to find the tour between particles such that each particle is visited once, and the total travelled distance is minimum (Gutin and Punnen, 2007; Diaby and Karwan, 2016). We used the Google OR-Tools, an open-source software for COP implemented in Python (Perron and Furnon, 2019). Given an arbitrary starting particle endpoint, the algorithm provides the explicit WHPA border representation which are the connected endpoints corresponding to the vertices of the closed curve. Applying TSP to each simulation, WHPAs are then stored in the $(n \times b \times 2)$-matrix $\mathbf{E}'$, the second dimension of $\mathbf{E}'$ being sorted. To the authors knowledge, it is the first time that TSP is used to determine WHPA's delineation. A WHPA can also be represented by a variable $\varphi$ over the grid such that $\varphi = 1$ for cells contained within the polygon defined by the solution of the TSP and $\varphi = 0$ for outer cells, defining a binary matrix. This binary WHPA representation corresponds to our target. Because PCA implicitly minimizes the least-square error of the distance between data points and their projections in the PC subspace, it is not appropriate for binary-valued data. To apply dimension reduction on the target, a suitable operator must be chosen to transform this Boolean matrix to a real-valued one, smoothness being a desirable property. Yin et al. (2020) demonstrated how to successfully apply PCA to a 2-dimensional discrete lithology model using the signed distance function (SDF) before dimension reduction. This approach is also adopted in this work. In two spatial dimensions, let $\mathcal{H}$ be the region of interest and $\mathrm{d}(\ )$ a Euclidean distance function defined as

$$\mathrm{d}(\vec{\mathbf{p}}) = \min(|\vec{\mathbf{p}} - \vec{\mathbf{p}}_b|) \ \forall \ \vec{\mathbf{p}}_b \in \partial\mathcal{H}, \qquad 8$$

entailing that $\mathrm{d}(\vec{\mathbf{p}}) = 0$ on the boundary where $\vec{\mathbf{p}} \in \partial\mathcal{H}$. A SDF is an implicit function $\psi$ with $|\psi(\vec{\mathbf{p}})| = \mathrm{d}(\vec{\mathbf{p}})$ for all points $\vec{\mathbf{p}} \in \mathbb{R}^2$. Hence, $\psi(\vec{\mathbf{p}}) = \mathrm{d}(\vec{\mathbf{p}}) = 0 \ \forall \ \vec{\mathbf{p}} \in \partial\mathcal{H}$, $\psi(\vec{\mathbf{p}}) = \mathrm{d}(\vec{\mathbf{p}}) \ \forall \ \vec{\mathbf{p}} \in \mathcal{H}^-$ (interior region) and $\psi(\vec{\mathbf{p}}) = -\mathrm{d}(\vec{\mathbf{p}}) \ \forall \ \vec{\mathbf{p}} \in \mathcal{H}^+$ (exterior region). In this case, the one-dimensional 0-contour of $\psi$ is chosen to represent the WHPA delineation that separates $\mathbb{R}^2$ into two separate subdomains with nonzero areas. An implicit representation means that the WHPA interface is represented as a one-dimensional isocontour of the higher-dimensional SDF, and not explicitly by all the points defining the curve.

Signed distance functions have the property

$$|\nabla \psi| = 1, \qquad (9)$$

$$\text{with } \nabla \psi = \left(\frac{\partial \psi}{\partial x}, \frac{\partial \psi}{\partial y}\right). \qquad (10)$$

The gradient $\nabla \psi$ is perpendicular to the isocontours of $\psi$. $|\psi(\vec{p})|$ gives the shortest distance from all points $\vec{p} \in \mathbb{R}^2$ to level sets of $\psi$ (Osher and Fedkiw, 2003). A fast numerical method to approximate SDF's is the Fast Marching Method (FFM), a scheme to solve the Eikonal equation (9) (Sethian, 1996). The FFM is implemented in the Python module scikit-fmm, which only works for regular Cartesian grids. The implicit representation is thus discretized into a uniform grid, so that the approximations errors are the same in both directions. Given $l_x$ and $l_y$ the total length of the computational grid in the x and y directions respectively, by choosing an adequate dimension of the cells in both axes, $\Delta x$ and $\Delta y$ ($\Delta x = \Delta y$ in a uniform grid), then the target matrix **V** has the shape $\left(n \times \frac{l_y}{\Delta y} \times \frac{l_x}{\Delta x}\right)$. Let $rows = \frac{l_y}{\Delta y}$ and $columns = \frac{l_x}{\Delta x}$. At this stage, the target is now represented as a closed surface whose interior and exterior regions are clearly defined by the $(n \times rows \times columns)$-shaped binary matrix **V**. The signed distance (SD) algorithm takes each sample of **V** as input array and computes their SD field. After completion, individual samples are the real, smooth $(rows \times columns)$ SD images, each pixel being a single feature. To apply dimension reduction, features need to be concatenated for each sample in the updated $(n \times rows \cdot columns)$-matrix **V'**. After application of PCA, the transformed pixel values lie in the PC's subspace, stored in $\mathbf{H}_\rho$ of shape $(n \times \rho)$, with $\rho = min(n, rows \cdot columns)$. To effectively reduce the target dimensions, the final step is to select the appropriate PC's number $1 \leq \upsilon < \rho$ to keep in the $(n \times \upsilon)$ target training set matrix

$$\mathbf{H}_\upsilon = \{\mathbf{h}_{i,j} \,|\, 1 \leq i \leq n;\ 1 \leq j \leq \upsilon\}. \qquad (11)$$

As in the predictor pre-processing step, PCA is only performed on the training set and the computed PCA coefficients are used to project the test target to the PC space. Note that it is only possible in the synthetic case, since in real conditions the true target would be unknown.

### *2.2.2. Training*

After pre-processing, a multivariate correlation between predictor $\mathbf{D}_\delta$ and target $\mathbf{H}_\upsilon$ is established by applying the CCA algorithm to train the model. The number of components $\eta$ is set to $min(\delta, \upsilon)$, i.e., the maximum possible number (Meloun and Militký, 2012). The resulting pairs of canonical variates (CV's) are stored in the $(n \times \eta)$ matrices

$$\mathbf{D}_\eta^c = \{\mathbf{d}_{i,1}^c, \mathbf{d}_{i,2}^c \ldots \mathbf{d}_{i,\eta-1}^c, \mathbf{d}_{i,\eta}^c |\ 1 \leq i \leq n\}, \qquad (12)$$

$$\mathbf{H}_\eta^c = \{\mathbf{h}_{i,1}^c, \mathbf{h}_{i,2}^c \ldots \mathbf{h}_{i,\eta-1}^c, \mathbf{h}_{i,\eta}^c |\ 1 \leq i \leq n\}. \qquad (13)$$

Normality is ensured in $\mathbf{D}_\eta^c$ and $\mathbf{H}_\eta^c$ by applying a Yeo-Johnson transform to each row (Yeo, 2000).

### *2.2.3. Regression*

Consider the predictor $\mathbf{d}_*$, a set of breakthrough curves that is not used in the training set. $\mathbf{d}_*$ goes through the same processing steps as the examples used to train the model, and the canonical weights computed at the training step are used to project the processed $\mathbf{d}_*$ to the canonical space.

This data is then used to infer $p(\mathbf{h}_*^c|\mathbf{d}_*^c)$, the posterior probability distribution in canonical space of the unknown target in light of $\mathbf{d}_*^c$ being observed, according to (1).

If linear correlation exists between the first CV pairs, MG inference can be performed to estimate the posterior multivariate normal distribution $p(\mathbf{h}_*^c|\mathbf{d}_*^c)$, i.e., infer its mean vector $\mathbf{m}_\eta^c$ of size $\eta$ (Eq. 3) and the $(\eta \times \eta)$ positive-definite covariance matrix $\mathbf{C}_\eta^c$ (Eq. 4).

### 2.2.4. Sampling

Samples are drawn from $p(\mathbf{h}_*^c|\mathbf{d}_*^c)$ to predict the distribution of the unknown WHPA given the trained regression model and the observed value. A number of samples $\zeta$ is chosen to allow proper graphical interpretation and uncertainty quantification. Each sample of the resulting $(\zeta \times \eta)$ matrix

$$\mathbf{H}_{*\eta}^c = \{\tilde{\mathbf{h}}_{*i,j}^c \mid 1 \leq i \leq \zeta;\ 1 \leq j \leq \eta\} \qquad 14$$

is backtransformed and reshaped to the corresponding $(\zeta \times rows \times columns)$ SD field, to obtain

$$\mathbf{H}_{*SD} = \{\tilde{\mathbf{h}}_{*i,j,k} \mid 1 \leq i \leq \zeta;\ 1 \leq j \leq rows;\ 1 \leq k \leq columns\}. \qquad 15$$

### 2.3. Experimental design

In the design step of an experiment, the observed data set is not known yet. Data sources can be placed virtually anywhere on the grid and the actual data could have any value within the prior data space. In this section, we demonstrate how BEL can be used to quantify the amount of information delivered by each possible data source, in order to make the optimal choice for the injection wells locations. To limit computational burden, we only test a finite number of predefined well locations, but the framework can be extended to all possible spatial configurations involving one or more wells on a given grid. For any selected configuration, our approach requires simulating the tracing experiments for every sample drawn from the prior model distribution.

In order to identify highly informative data sets based on their location, some data-utility function must be maximized or minimized.

### 2.3.1. Informative data sets identification and data utility function

To assess which data source is the most informative among a set of $s$ possible locations, we estimate which data source yields the largest reduction of uncertainty from the prior to the posterior. However, quantifying the uncertainty variation is not straightforward as WHPA's are high-dimensional. A meaningful approach to assess the uncertainty reduction for each of the $s$ data sources, given $\zeta$ drawn samples, is to compare the true SD images with the generated ones for each test model, such that

$$\boldsymbol{\theta} = \left\{\sum_{j=1}^{\zeta} \mathrm{H}(\mathbf{h}_*^\eta, \tilde{\mathbf{h}}_{*i,j}) \mid 1 \leq i \leq \lambda\right\} \qquad 16$$

is the vector of length $\lambda$ whose entries are the summed computed discrepancies over each posterior target sample for each data source and a given true (test) model or image. If $\mathbf{h}_*$ is the true $(rows \times columns)$ SD image corresponding to the predictor $\mathbf{d}_{*i}$, i.e., the single breakthrough curve of tracer $i$, then $\mathbf{h}_*^\eta$ is the backtransformed version of $\mathbf{h}_*$ using $\eta$ components. $\tilde{\mathbf{h}}_{*i,j}$ contains the $\zeta$ drawn samples using a unique data source $i$. Each entry of $\tilde{\mathbf{h}}_{*i,j}$ has the same dimension as $\mathbf{h}_*$. $\mathrm{H}(\ )$ is the operator that measures the similarity between two images. The distribution of $\boldsymbol{\theta}$ can then be examined to identify the data source index having the minimum or maximum range, depending on the measure of similarity $\mathrm{H}(\ )$, indicating the most informative source.

Dubuisson et al. (2004) and Scheidt et al. (2018) recommend using the Modified Hausdorff Distance (MHD) as the data-utility function. It is particularly adapted to the problem at hand, since it implies measuring the distances between points forming the edges of the features of interest in the images. In this case, WHPA's are implicitly represented by the 0-contours edges of the SD images, whose coordinates can be extracted. Let $\mathcal{D} \in \mathbb{R}^{m \times w}$ be the pairwise Euclidean distance matrix between two vectors $\vec{p}_1 \in \mathbb{R}^{m \times 2}$ and $\vec{p}_2 \in \mathbb{R}^{w \times 2}$. The MHD between $\vec{p}_1$ and $\vec{p}_2$ is

$$\text{MHD} = \max\left[\overline{\min_i[\mathcal{D}_{ij}]}, \overline{\min_j[\mathcal{D}_{ij}]}\right] \in \mathbb{R}, \qquad 17$$

$$\min_i[\mathcal{D}_{ij}] \in \mathbb{R}^w,$$

$$\min_j[\mathcal{D}_{ij}] \in \mathbb{R}^m,$$

where $1 \leq i \leq m; 1 \leq j \leq w$ are respectively the row and column index of $\mathcal{D}$. The overbar denotes the mean operator. $\mathcal{D}_{ij}$ is the Euclidean distance between the $i^{th}$ point of $\vec{p}_1$ and the $j^{th}$ point of $\vec{p}_2$. The max operator is used in (17) because $\mathcal{D}$ is not symmetric. The MHD is robust and is monotonically increasing as the dissimilarity between two contours increase, thus a smaller value of $\boldsymbol{\theta}$ indicates the most similarity between images (Dubuisson et al., 2004).

Another candidate for the data-utility function is the Structural Similarity (SSIM) index, a metric that measures the similarity between two continuous images (Wang et al., 2004). It is symmetric ($\text{SSIM}(\text{im}_1, \text{im}_2) = \text{SSIM}(\text{im}_2, \text{im}_1)$) and bounded from 0 to 1 ($\text{SSIM}(\text{im}_1, \text{im}_2) = 1$ if $\text{im}_1 = \text{im}_2$) for two images $\text{im}_1, \text{im}_2$ of the same region in space. This index can be directly computed using the SD images of the reference (true) target and the sampled ones. It takes the form

$$\text{SSIM}(\text{im}_1, \text{im}_2) = F(L(\text{im}_1, \text{im}_2), C(\text{im}_1, \text{im}_2), S(\text{im}_1, \text{im}_2)). \qquad 18$$

The three terms $L, C, S$ are the luminance, contrast and structures components respectively. $F$ is the combination function. For more details, see Wang et al. (2004).

However, (16) provides the uncertainty range for one observation set, i.e., a single test example. In experimental design, the actual data is unknown and could take any value in the prior range. Assumptions on the prior model distribution are made by assigning the $n$ hydraulic conductivity $\mathbf{K}$ fields to the computational grid, thus a single performance measure is not enough to define the value of information of each injection well. A common practice in machine learning is to use 80% of the total dataset for training and hold out the remaining 20% as test set (Géron, 2019). Therefore, we combine the uncertainty range for each data source for a series of possible observations

$$\boldsymbol{\Theta} = \{\boldsymbol{\theta}_i \mid 1 \leq i \leq n_{20\%}\} \qquad 19$$

is the matrix of shape $(\lambda \times n_{20\%})$ containing the summed MHD of each data source for each member of the test set of size $n_{20\%}$. In order to fully exploit the available dataset and to control the possible sensitivity of the results to the chosen training and test sets, k-fold validation is performed. The k-fold procedure randomly splits the available dataset into $k$ distinct subsets called folds. Then, the model is trained and the error measured $\boldsymbol{\Theta}$ evaluated $k$ times, picking a different fold for testing, and training on the remaining $k-1$ folds (Géron, 2019). In the end, a number $k$ of $\boldsymbol{\Theta}$ matrices are obtained, one for each k-fold combination.

In this work, we identify the most informative data sources by looking at the statistical summary of their distribution using boxplots. If the interquartile range (IQR) and/or the median for each data source between the $k$ $\boldsymbol{\Theta}$ matrices differ, i.e., are inconsistent with each another, then that means that the dataset used is not large enough. This assessment is a visual task depending on the experimenter, although one could automate this process by using for example the mode of each distribution, the IQR, or a combination of criteria as a proxy to evaluate informative data sources.

In contrast to other approaches used for experimental design, the advantage of BEL is that the training step allows to readily calculate the posterior distribution for any observed data. Therefore, solving the optimal design problem only requires running the forward model for the training set and estimating the information content of different source is obtained at little additional costs. It also allows to define several data utility function on the basis of the posterior prediction ranges, without any restriction. We therefore deem BEL to be particularly appropriate for experimental design.

### 3. Application

We calculate the WHPA from tracing experiments in a single layer aquifer. Following BEL, we sample the prior distribution of the model parameters by creating $n$ models $\{\mathbf{M}_1, \mathbf{M}_2 \ldots \mathbf{M}_{n-1}, \mathbf{M}_n\}$ and simulate both the WHPA using particle backtracking and BC's using solute transport. We thus solve $n$ solute transport problems and $n$ backtracking problems.

A 2-dimensional grid is considered for the experiments and the MODFLOW-2005 code is used for groundwater flow modelling (Harbaugh et al., 2017). The whole code is built with the Python-written FloPy toolbox (Bakker et al., 2016), including support for MODFLOW-2005, MT3DMS-USGS and MODPATH 7.

The computation of a single pair of forward modellings (transport and backtracking) on a 2.3 GHz 8-Core Intel Core i9 processor was timed to 37 minutes.

#### *3.1. Groundwater model*

The different steps of the experiment (flow, transport and particle tracking) depend on the unknown hydraulic conductivity $\mathbf{K}$ field. Using SGEMS (Remy et al., 2009), $n$ samples from the prior distribution are generated by Sequential Gaussian Simulation (SGS) (Goovaerts, 1997). The spatial correlation of the $\log_{10} \mathbf{K}$ field is defined by a variogram model (see Table 1). In this paper, only the mean value of the $\log_{10} \mathbf{K}$ field is considered unknown, but this could be extended to any other parameters without loss of generality (see Hermans et al., 2018, 2019).

| Parameter | Value |
|:---:|:---:|
| Grid x-extent $(m)$ | 1500 |
| Grid y-extent $(m)$ | 1000 |
| Grid z-extent $(m)$ | 10 |
| $n_{row}$ | 157 |
| $n_{col}$ | 207 |
| $n_{lay}$ | 1 |
| $SP_1$ $(days, \text{steady state})$ | 1 |
| $SP_2$ $(days, \text{transient})$ | 0.08 |
| $SP_3$ $(days, \text{transient})$ | 100 |
| Pumping well rate $\left(\frac{m^3}{d}\right) [SP_1, SP_2, SP_3]$ | -1000 |
| Injection wells rate $\left(\frac{m^3}{d}\right) [SP_2]$ | 24 |
| Tracers mass loading $\left(\frac{kg}{d}\right)$ | 1.5 |
| $S_s$ $(m^{-1})$ | $10^{-4}$ |
| $S_y$ (-) | 0.25 |
| $\alpha_L$ (m) | 3 |
| K mean range $\left(\frac{m}{d}\right)$ | [25, 100] |
| $\log_{10}$K standard deviation $\left(\frac{m}{d}\right)$ | 0.4 |
| Kriging type | Simple |
| Nugget effect | 0 |
| Structure | Spherical |
| Range (m) [min, max] | [25, 100] |

*Table 1. Model parameters*

A number *s* of injection wells are placed around a pumping well. Their role is to inject individual tracers to model their transport and record their breakthrough curves (BC's) at the pumping well location. Several particles are artificially placed around the pumping well and their origin after a given amount of time is backtracked to delineate the corresponding WHPA.

The structured grid has dimensions of $1500\ m$, $1000\ m$ and $10\ m$ in the x, y, z axes respectively. The base cell dimension in the x, y axes is $10\ m$, incrementally refined around the pumping well ($1000\ m$, $500\ m$), down to $1\ m$ by $1\ m$ cells. The pumping rate is $1000\ \frac{m^3}{d}$. The dimensions $(n_{row}, n_{col}, n_{lay})$ of the grid are (157, 207, 1). The total flow simulation period is discretized into 3 stress-periods $(SP_1, SP_2, SP_3)$ during which pumping occurs at all time steps. $SP_1$ is steady state, to compute the hydraulic heads in pumping conditions. Injection occurs during the transient $SP_2$ which is subdivided into 300 time steps of 24 seconds each. $SP_3$ is transient, 100 days long, and is discretized into 100 time steps to model tracer transport from the injection wells and record their BC's at the pumping well location. The flow boundary conditions are a fixed head of $0\ m$ along the western boundary, and a fixed head of $-3\ m$ along the eastern boundary. They are kept constant across all stress-periods to create a negative gradient in the x direction. Note, however, that boundary condition uncertainty could also be included in the framework (Hermans et al., 2018, 2019). Six injection wells are placed around the pumping well (Figure 1). They inject individual tracers with mass loading of $1.5\ \frac{kg}{d}$, at the rate of $24\ \frac{m^3}{d}$ for two hours $(SP_2)$.

This aquifer layer is modeled as a confined aquifer. Its specific storage $\mathbf{S_s}$ and specific yield $\mathbf{S_y}$ are constant and set to $10^{-4} m^{-1}$ and 0.25 (no units) respectively. For transport modelling, the hybrid method of characteristics (HMOC) is used to solve for advection. The porosity is set to the $\mathbf{S_y}$ field and the longitudinal dispersion $\mathbf{\alpha_L}$ is set to $3\ m$.

For defining the WHPA in pumping conditions with particle-tracking, 144 particles are placed around the pumping well and their backward endpoints after 30 days are computed. The porosity value required for the particle-tracking algorithm is set to 0.25. The above parametrization is kept constant.

The $\log_{10}\mathbf{K}$ mean is randomly varied between 1.4 and 2, so that the $\mathbf{K}$ mean is ranging from 25 to $100\ \frac{m}{d}$, which are values commonly observed in alluvial aquifers (Dassargues, 2020). The standard deviation of $\log_{10}\mathbf{K}$ is set to $0.4\ \frac{m}{d}$ and remains constant. To ensure pumping and tracer injection, sufficiently high $\mathbf{K}$ values are fixed at all well locations, randomly chosen between 100 and $1000\ \frac{m}{d}$. The output of such a simulation is illustrated on Figure 1, and the resulting computed hydraulic heads on Figure 2.

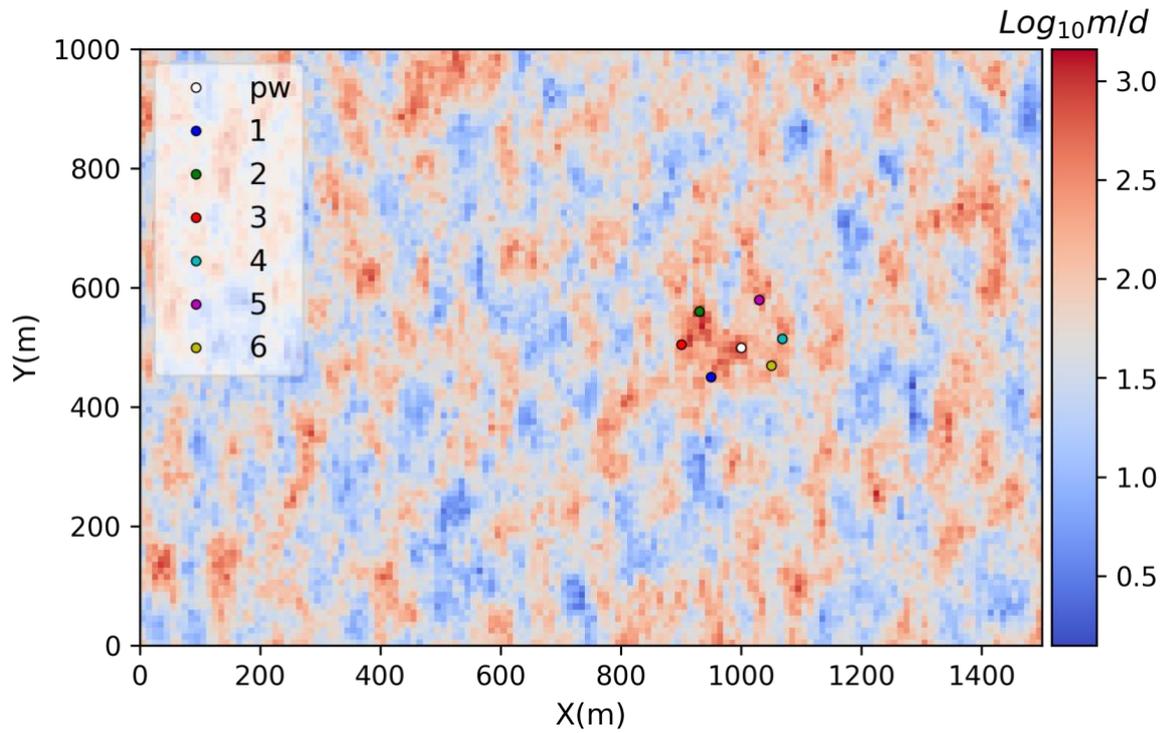

Figure 1. Hydraulic conductivity field in 10-logarithmic base. The pumping well (pw) is located at (x, y) = (1000m, 500m) and is surrounded by 6 injection wells.

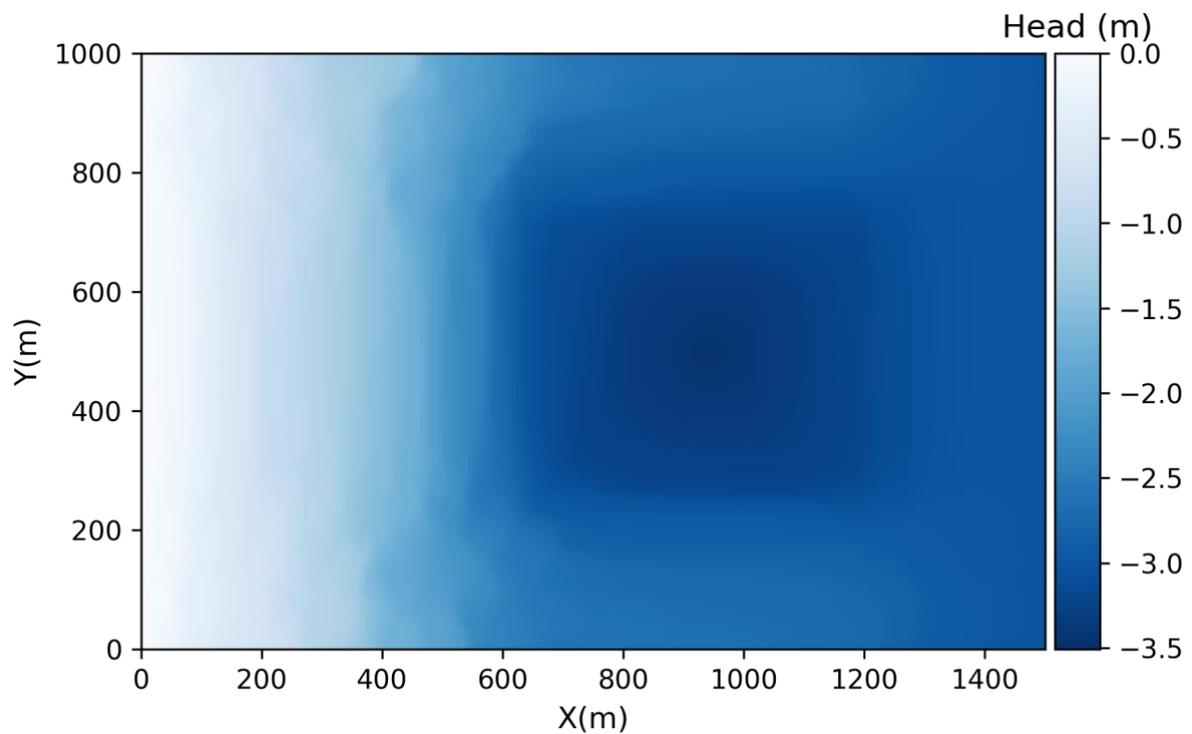

Figure 2. Flow solution. The direction of the natural gradient is from west to east.

### 3.2. WHPA prediction

This section illustrates how to predict a single WHPA using all sources of information, i.e., the 6 injection wells. We demonstrate that BEL can efficiently estimate the posterior distribution of the target WHPA. The total size of the dataset in this section is $n = 500$. The size of the training set is then $n_{80\%} = 400$. Such a relatively low training size is sufficient because the prediction is much simpler than the underlying model. Previous BEL applications have demonstrated that this order of magnitude is sufficient to provide accurate predictions (Hermans et al., 2016, 2018, 2019; Athens and Caers, 2019; Michel et al., 2020a, 2020b; Park and Caers, 2020; Yin et al., 2020). For demonstration purpose, 4 examples will be picked out of the remaining 100 samples to illustrate the prediction capabilities of the BEL framework. However, only one of them will be used to demonstrate the different steps, thus resulting in a test set of size 1 in the following.

#### *3.2.1. Pre-processing*
#### 3.2.1.1. Predictor

First, the BC's are interpolated through 200 pre-defined equidistant time-steps, which significantly reduces the dimensionality of the predictor (Figure 3). The 6 curves of each set are then concatenated into the predictor matrix of shape $(n_{80\%} \times \lambda \cdot k) = (400 \times 1200)$ and PCA is applied to obtain the square PC matrix $\mathbf{D}_{400}$. An adequate number $\delta$ of PC's must be chosen to truncate the columns of $\mathbf{D}_{400}$, to retain a sufficient amount of information while reducing the original dataset dimension. Setting $\delta$ to 50 allows to explain 99.87% of the variance of the data while reducing its size by a factor 4. The final product of the predictor's pre-processing is the $(400 \times 50)$-$\mathbf{D}_\delta$ matrix (5). The remaining 0.13% of variance is not considered useful for predicting the target. Figure 4A illustrates the full predictor for the chosen example (6 concatenated curves) and its reconstruction using the 50 firsts PC's. The back transformation recovers the original structure well, though slightly smoothing out the curves. The raw BC's of the chosen example have size $(6 \times 18444)$ (i.e., there are 18444 time steps for each curve) and the dimensionality of the original data is thus divided by a factor of 369 by the joint interpolation-PCA preprocessing, while preserving most of the information. As the number of PC increases, the range of coefficients for the model decreases (Figure 4C-D). This reflects the fact that a lower amount of information on the curves gets stored in them, which is additionally depicted on Figure 4E-F which illustrate the cumulative explained variance with the number of PC's. The PC's of the test predictor is compared to those of the training set on Figure 4C, illustrating that this test predictor is consistent with the prior.

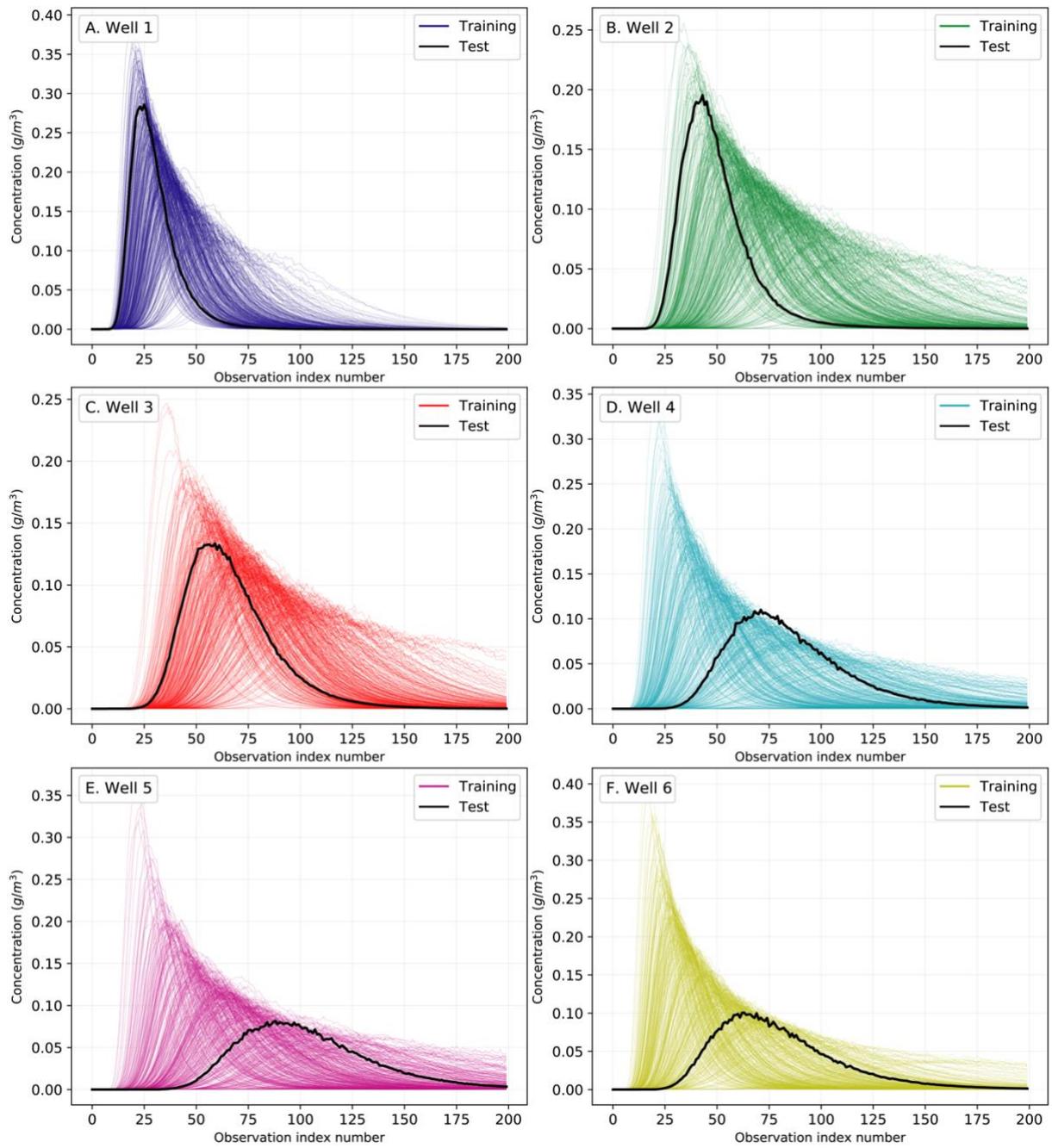

*Figure 3. Breakthrough curves of tracers from each injection well, for both training and test (single sample) sets. They are all discretized and interpolated in 200 steps.*

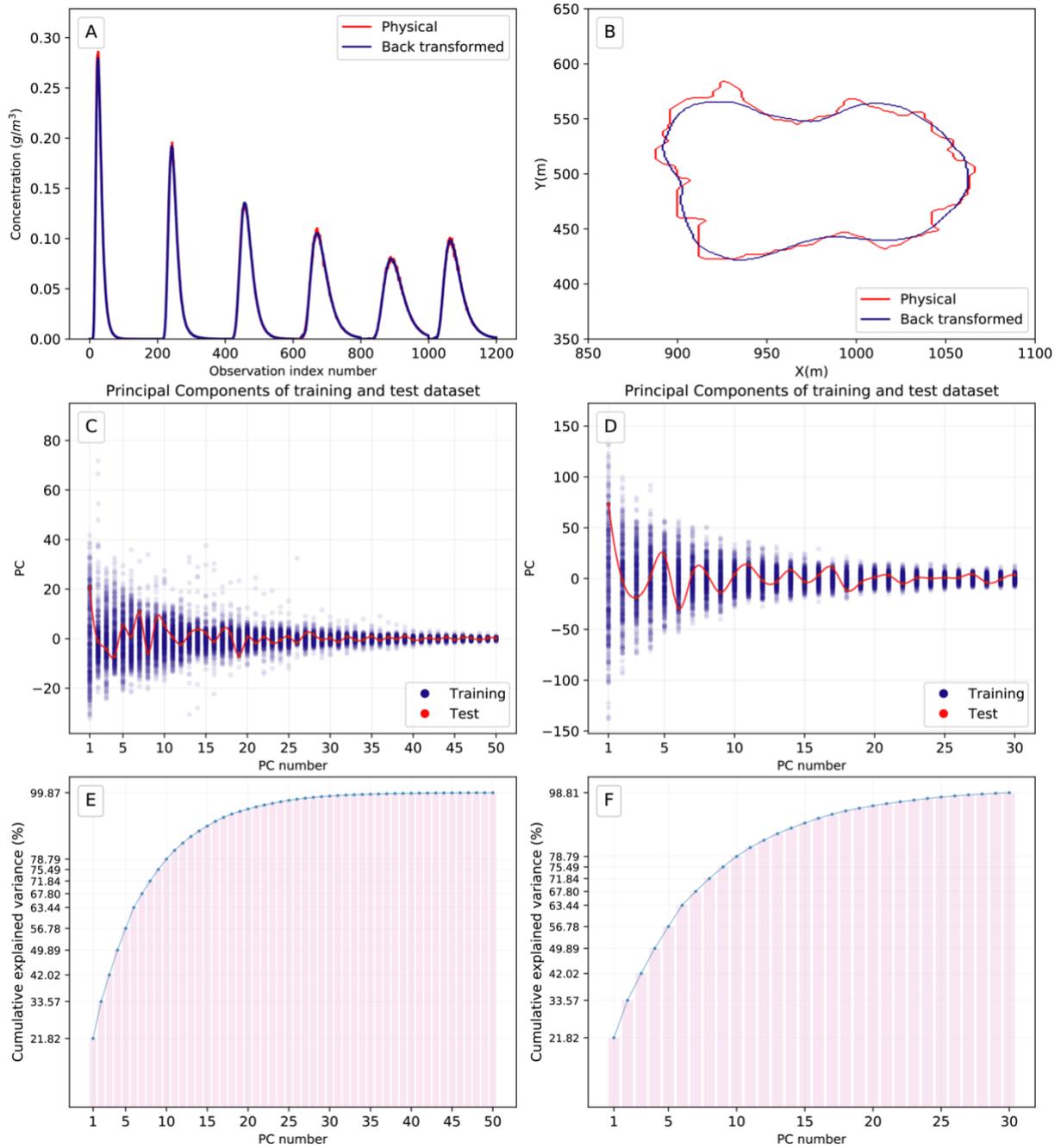

Figure 4. *A.* The full predictor of the *test* set is the concatenation of all breakthrough curves (black curves on Figure 3). PCA decomposition with 50 PC's allows to recover the original curves while smoothing out some noise present in the original dataset. *B.* Zero-contour of the original Signed Distance of the *test* target compared to the zero-contour of the Signed Distance of the *test* target projected then back-transformed with its 30 Principal Components. *C.* Principal Components of the predictor *training* set and projected *test* set. *D.* Principal Components of the Signed Distance of the target *training* set and the projected *test* set. *E.* Cumulative explained variance for the Principal Components of the predictor *training* set. *F.* Cumulative explained variance for the Principal Components of the Signed Distance of the target *training* set.

### 3.2.1.2. Target

The raw output of the backtracking simulation for the given, randomly chosen test example is illustrated on Figure 5A. The 144 2D-coordinates are scattered around their origin, the pumping well location. A few points are randomly selected and their index in the output file is displayed. The next step is to convert each item to an image using the SD algorithm. Since endpoints have limited travel extents, a focused area is defined to reduce the computational cost of the SD estimation and storage using: $(x_{min}, x_{max}) = (800\ m, 1150\ m)$ and $(y_{min}, y_{max}) = (300\ m, 700\ m)$. This subdomain is subdivided into $4 \cdot 4\ m^2$ cells, resulting in 100 rows and 87 columns for a total of 8700 cells. The cell dimensions are adapted to keep enough information on the WHPA without having to store a too heavy image. The WHPA delineations are now represented by the zero-contour of the binary-field. This Boolean matrix is then fed to the SD algorithm which computes the SD value of this isocontour for each cell of the domain, as illustrated on Figure 5B for the chosen test example. Figure 5C shows the 0-contours of the training and test WHPA's SD. Each SD image is then flattened and PCA is applied on the resulting $(400 \times 8700)$-matrix. The number of PC's, $v = 30$, is chosen based on the reconstruction of the WHPA image as illustrated in Figure 4B. Because the WHPA is very complex as a result of the **K** distribution, it would be illusory to try to predict all the nooks that cannot be captured by the tracing experiment. Instead, the slightly smoothed-out reconstructed WHPA is a good candidate for prediction. By making these choices, the resulting target is the $(400 \times 30)$-$\mathbf{H}_v$ matrix (11). The content of $\mathbf{H}_v$ is displayed on Figure 4D for the training and test set, and the cumulative explained variance on Figure 4F, reaching the amount of 98.81% for the last PC. Note that the remaining 1.19% that are not predicted could be added as random noise after prediction (Park and Caers, 2020). However, since the final purpose is experimental design, there is no interest in adding the same uncertainty component to all samples.

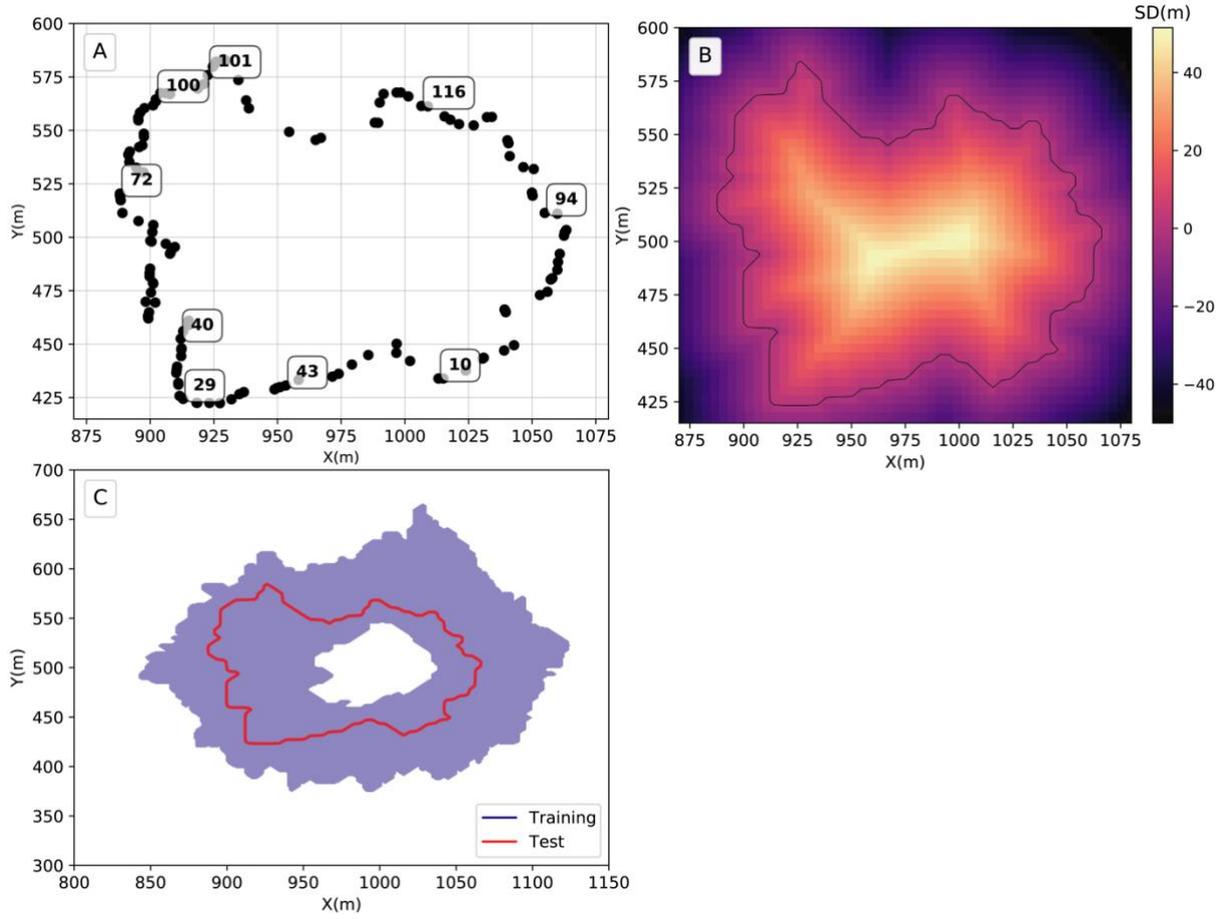

*Figure 5. **A**. Raw representation of the chosen **test** target. A few points indexes are randomly highlighted among the total 144 particles to illustrate the meaningless ordering of the endpoints output. **B**. Implicit representation of the **test** WHPA on a discretized grid. The WHPA delineation corresponds to the zero-contour of the SD field which is computed for each cell as the closest distance from their center to the boundary. **C**. Target **training** set and chosen **test** example in physical space.*

### 3.2.2. Training

CCA is applied between $\mathbf{D}_{50}$ and $\mathbf{H}_{30}$. The maximum number of CV's thus equals 30. The resulting Canonical Pairs matrices are $\mathbf{D}_{30}^c$ (matrix 12) and $\mathbf{H}_{30}^c$ (matrix 13). The first 3 CV pairs are illustrated on Figure 6A to Figure 6C, showing high linear correlations whose strengths are reflected by the Canonical Correlation Coefficient $r$ (Meloun and Militký, 2012), whose decreases with the number of CV's is shown on Figure 6D. For each pair, a KDE is performed, and its density plotted behind the CV's point cloud. It is here done purely for visualization of the joint probability distributions, but it could also be used for sampling as an alternative to linear regression in case of non-linearity or non-Gaussianity (Michel et al., 2020a), as illustrated on the y marginal plots of Figure 6A to Figure 6C.

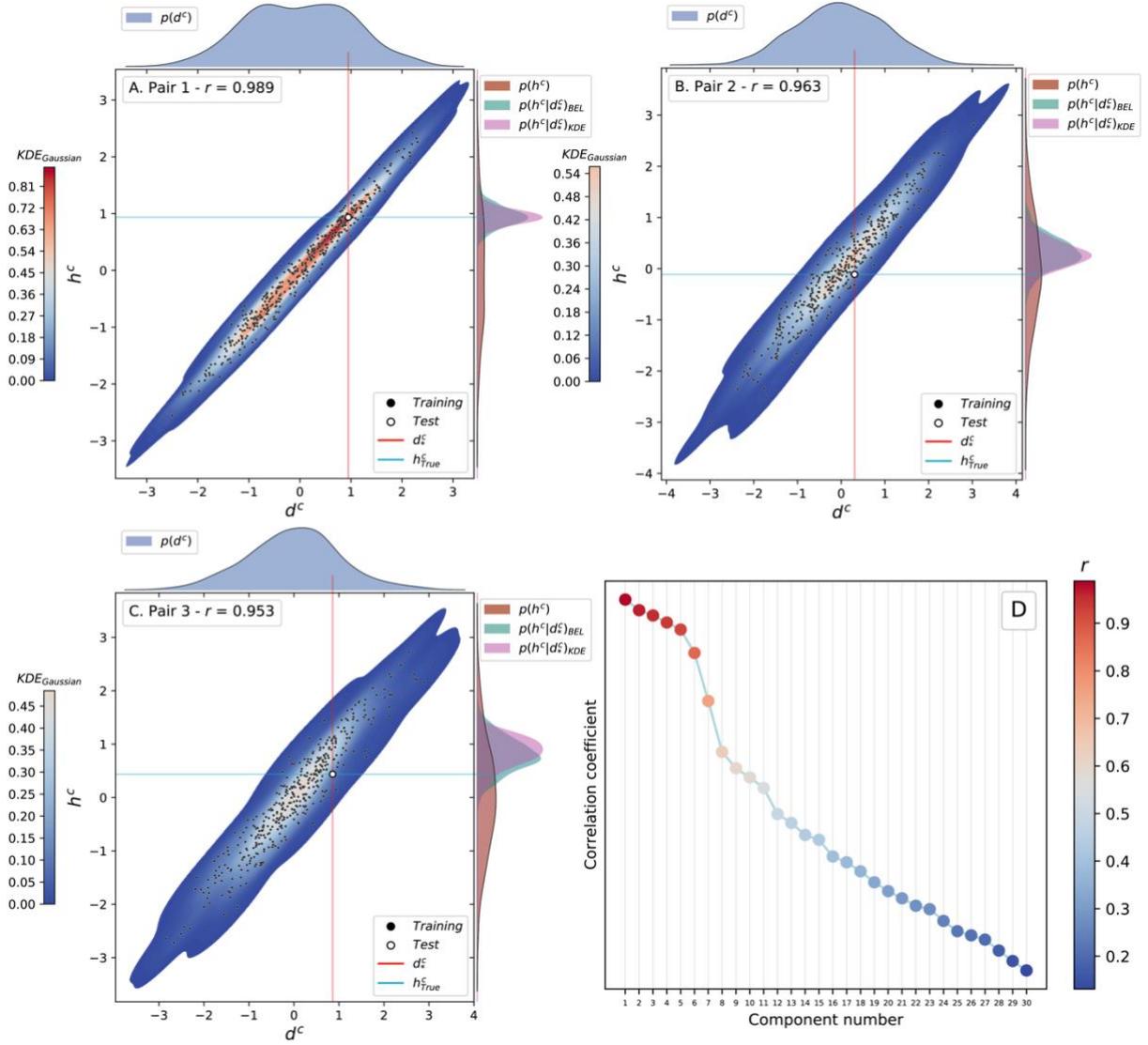

Figure 6. *A, B, C.* Canonical Variates bivariate distribution plots for the 4 firsts pairs of the *training* set, and the projection in the canonical space of the selected *test* predictor and associated *test* target (see notches). The posterior distribution of $\mathbf{h}^c$ computed according to BEL and KDE can be compared on the y marginal plot. *D.* Decrease of the Canonical Correlation Coefficient r with the number of CV pairs for the *training* set.

### 3.2.3. Regression

The test predictor depicted in Figure 4C is projected to the canonical space as illustrated on Figure 6A-C. MG inference is then applied to infer the posterior covariance (Eq. 3) and mean (Eq. 4).

### 3.2.4. Sampling

As explained in section 2.1, samples are drawn from the MG described by its two firsts moments and backtransformed to the original space. **Error! Reference source not found.**A depicts 400 WHPA's (arbitrary number, chosen for proper visual comparison) sampled from the inferred posterior distribution of the chosen example. The 400 samples englobe the true prediction, and the uncertainty reduction is visible by comparing them with the footprint of the prior prediction range. The uncertainty reduction around injection wells is visible as knots where the WHPA distribution tightens. The methodology output is illustrated for three other (randomly chosen) test examples (**Error! Reference source not found.**B-D), using the same training set. In each case, BEL successfully identifies the prediction with its uncertainty range. Of course, given the limited amount of data, the range of uncertainty in the posterior remains important.

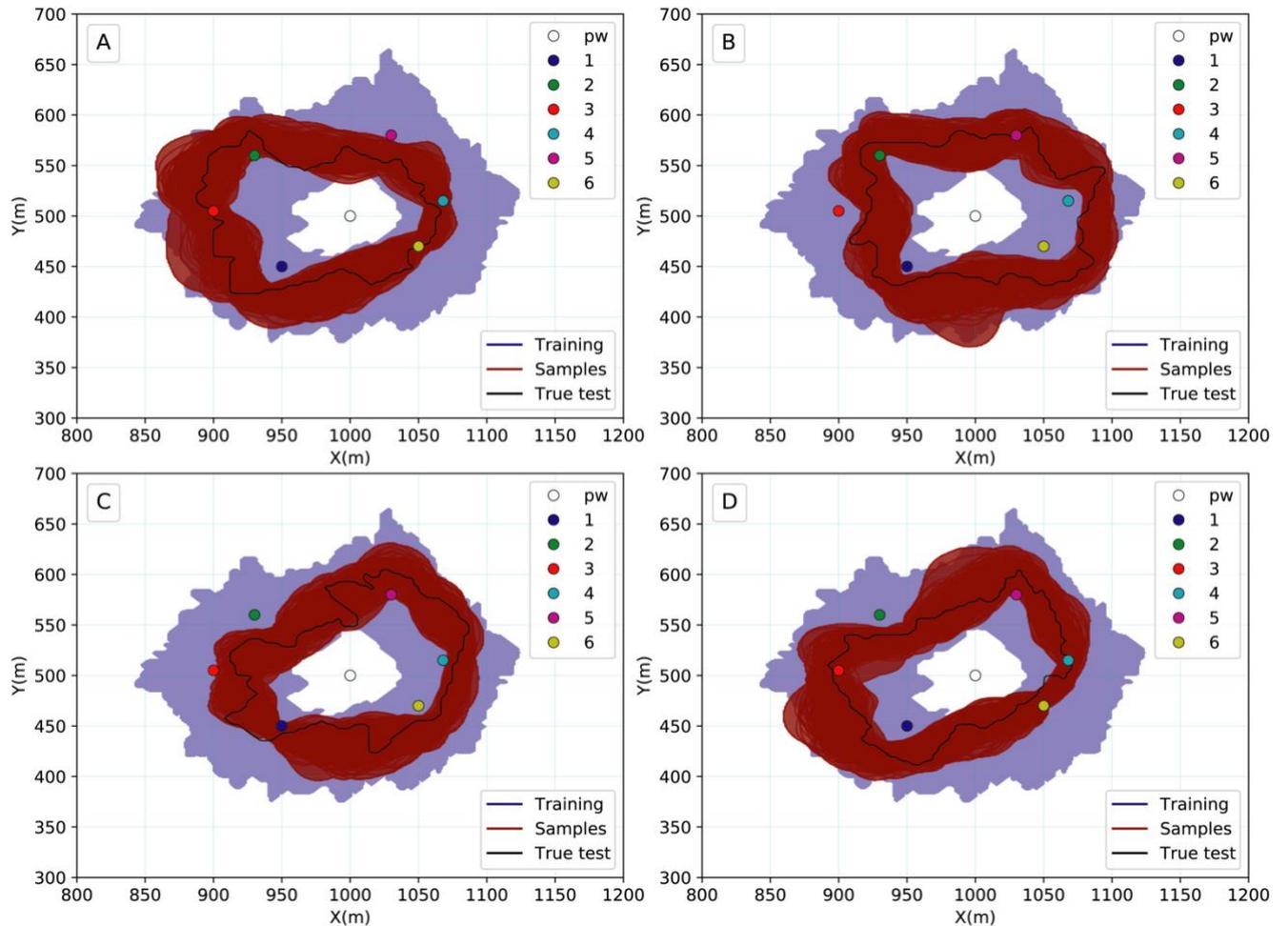

*Figure 7. BEL-derived posterior predictions for four different tests WHPA's. Subfigure A displays the chosen test example associated with Figures 3-6.*

### 3.2.5. Influence of the training set size

In Figure 8, we validate the choice of a training set of 400 models to compute the posterior. The figure shows how the average SSIM index values vary with different training set sizes, from 125 to 900, each computed with 400 samples of the posterior distribution of randomly chosen targets (each line corresponding to a different WHPA). Only 20 different targets are displayed for visualization purposes. It shows that from 400 models, the average metric values stabilize. Therefore, a training set of such size is appropriate to predict the posterior distribution of the target for a broad range of different WHPA's. This training set size will also be used for experimental design.

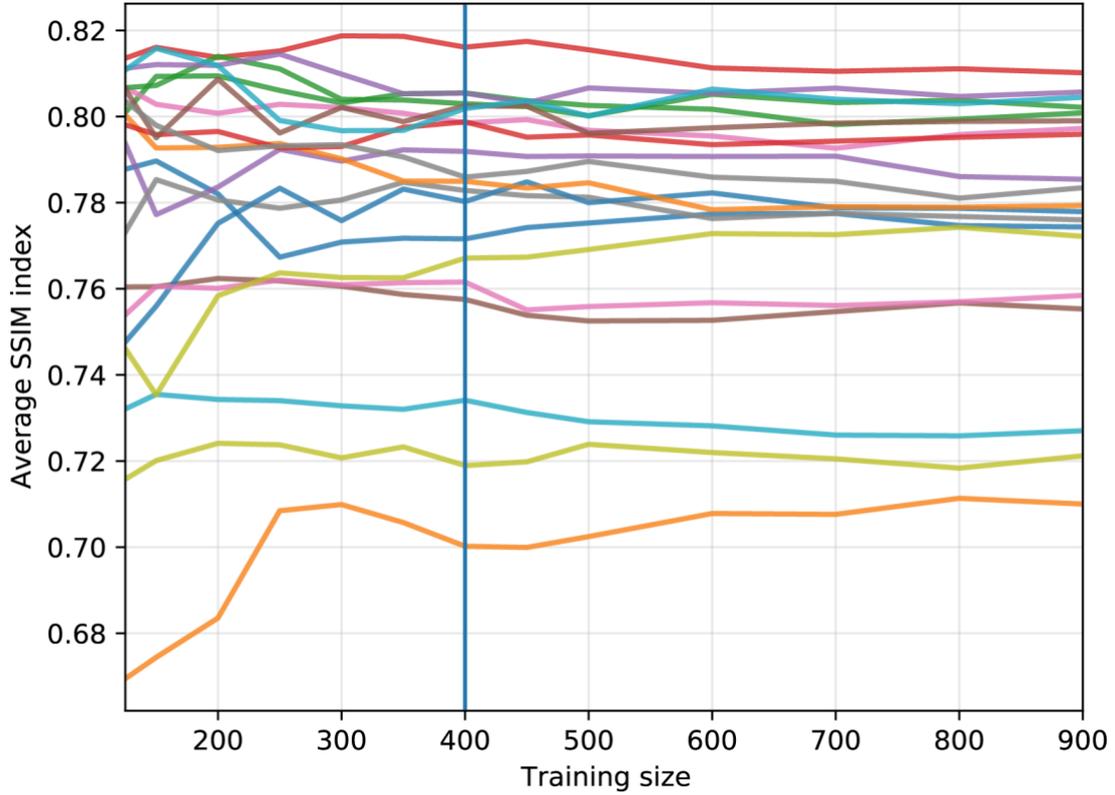

Figure 8. Effect of the training set size (125 to 900) on the average SSIM index for 20 different targets. Each line corresponds to a single target being predicted with training sets of increasing size. A stabilization of the average metric value occurs at around $n_{training} = 400$. The maximum value of the SSIM index is 1 for exactly similar images.

## 4. Experimental design

In this section, the optimum location of a single well is estimated by making a single prediction on each individual source, i.e., $\lambda = 1$ with subsets [1], [2], [3], [4], [5], [6]. The same number of training samples are used as in the regression case (section 3.2). $n_{training} = n_{80\%} = 400$ and $n_{test} = n_{20\%} = 100$.

Figure 9A-F illustrate the effect of using a single data source at a time to predict the example WHPA. As expected, the uncertainty is reduced around the sources and englobes the training set in further regions of the grid. In order to find out which injection well location is the most informative, sampling is carried out in the same fashion for the $n_{20\%} = 100$ test set (not shown). For each one of them, 400 samples are drawn for each well and the MHD's and SSIM index between drawn and "true" target are computed. The MHD and SSIM index of each sample are summed over the 100 test examples and the resulting boxplots are shown in Figure 10A-B, respectively. The chosen convention is that a higher metric value implies further distance from the true target. To enforce this, the opposite of the SSIM index value is used. The MHD and SSIM index metric are standardized by removing their mean and scaling to unit variance. Both metrics then provide very similar results: from Figure 10A-B, it appears that, globally, wells downstream the pumping well (Figure 1), i.e., tracers going against the natural gradient (Figure 2) contain a larger amount of information (wells 4, 5 and 6), as indicated by their IQR bounds lower in magnitude compared to the IQR of the upstream wells 1, 2 and 3. As an illustration, the boxplot is shown for the single test example considered previously. Figure 10C–D

identify well 5 as the most informative for this particular case. This illustrates that the results for one individual data set do not allow to identify the most informative well globally.

In order to validate the conclusions made on the most informative wells as inferred from Figure 10A-B, k-fold cross-validation is performed on the whole set of $n$=500 considered samples. Five splits are chosen, so that the set of samples is successively cut into 400 training and 100 test samples. Figure 11 illustrates that a set of 100 test samples is inappropriate to perform experimental design in the considered case. Indeed, the boxplots corresponding to the 5 different splits from Figure 11A to 11E are inconsistent with each another. The 5-fold cross-validation procedure is thus successively repeated with larger datasets until the boxplots start to be consistent with each other across splits. It is shown in Figure 12 that a set of size $n$=1250 (successively cut into 1000 training samples and 250 test samples) produces consistent metrics across all 5 splits and validates the interpretation that the most informative data sources are the injection wells number 4, 5 and 6 (downstream wells). Upstream wells are consistently ranked lower in terms of information content, with well number 1 being the worst in all cases. On the other hand, the most informative data source can be inferred as well number 6, constantly showing low-bounds and narrow IQR, as opposed to well 4 showing broad IQR bounds in the 4$^{th}$ split (Figure 12D). Therefore, in this case, for consistent experimental design of WHPA prediction, we find that the test data set should contain at least 250 samples.

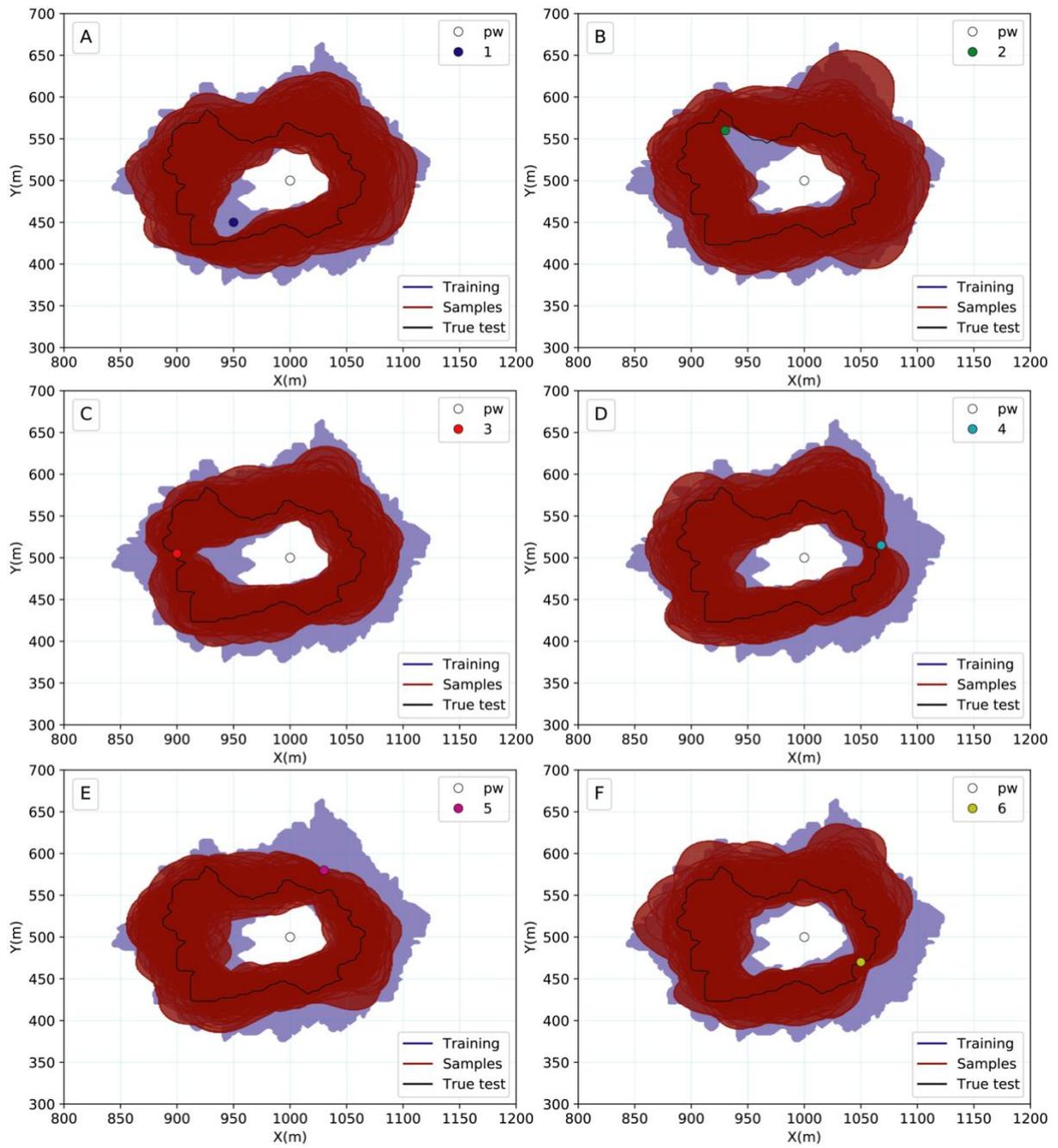

Figure 9. WHPA predictions for each individual well 1, 2, 3, 4, 5, 6 for the chosen test example.

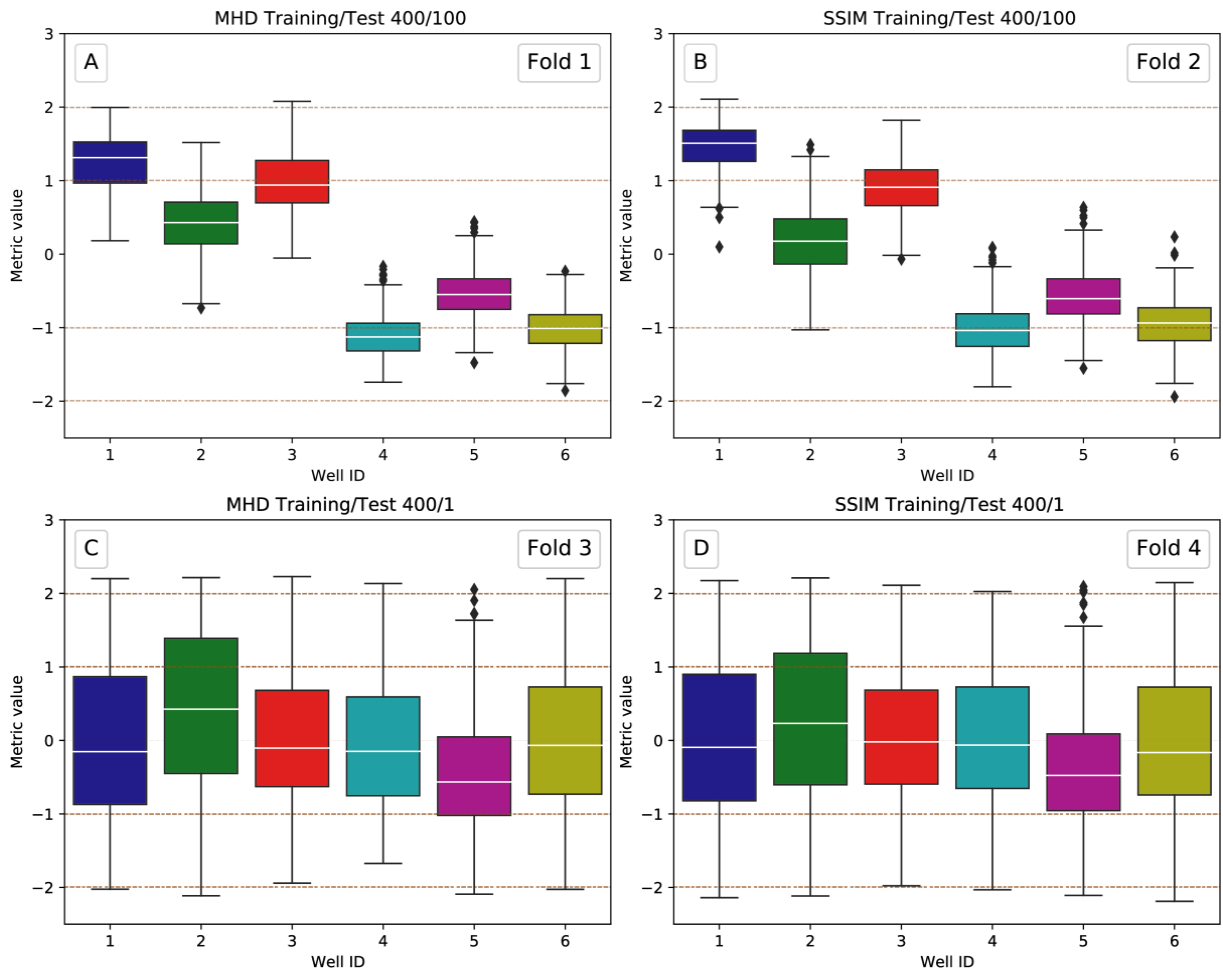

Figure 10. **A.** Boxplots of the standardized MHD distance for each well and the 100 test samples. **B.** Boxplots of the standardized SSIM index for each well and 100 test samples. **C.** Boxplots of the standardized MHD distance for each well and the single test sample. **D.** Boxplots of the standardized SSIM index for each well and the single test sample.

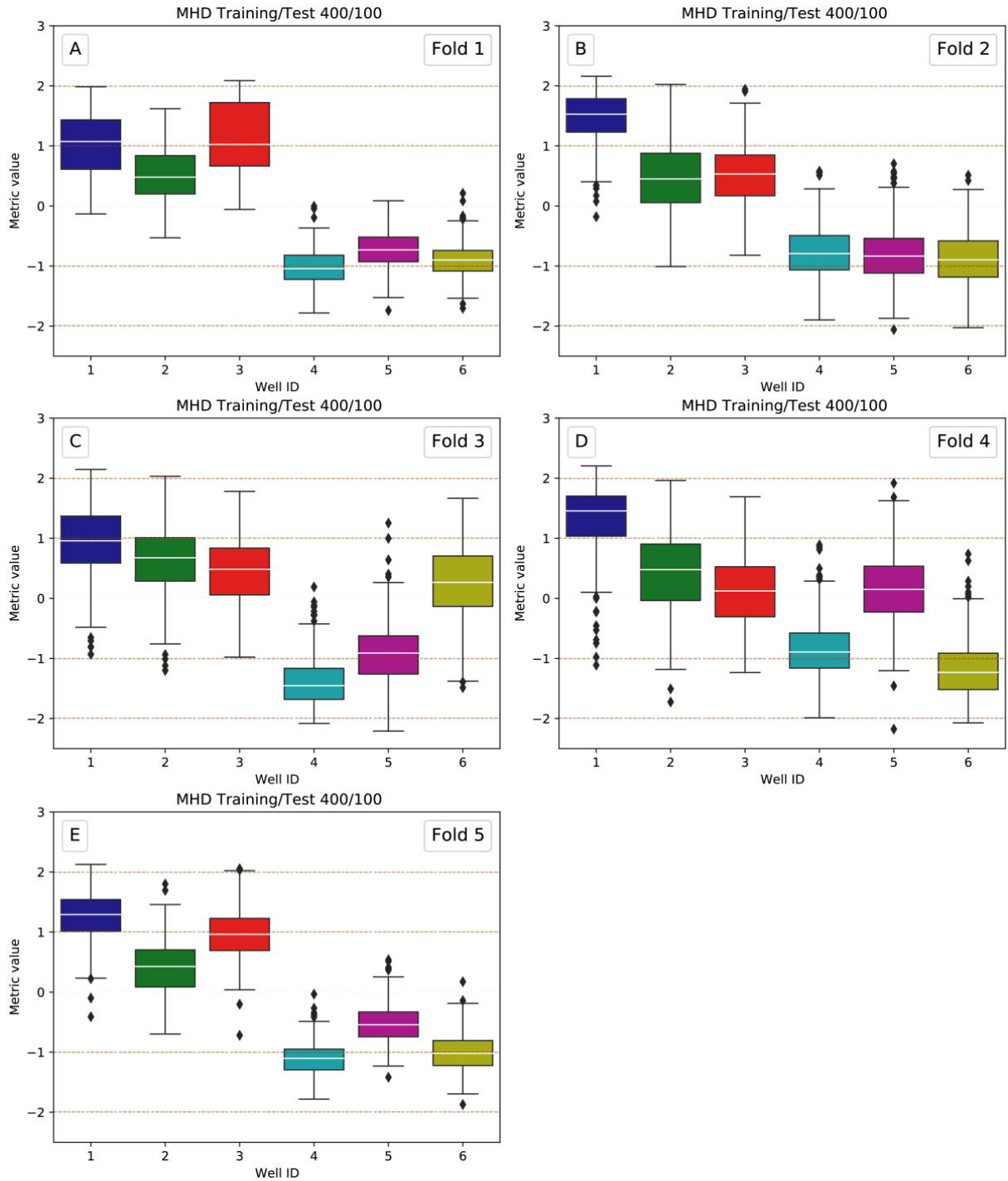

Figure 11. Boxplots of the standardized MHD distance for each well and the 5 successive k-fold for a training dataset of size **400** and **100** test samples. The boxplots for each well are inconsistent with each other across folds.

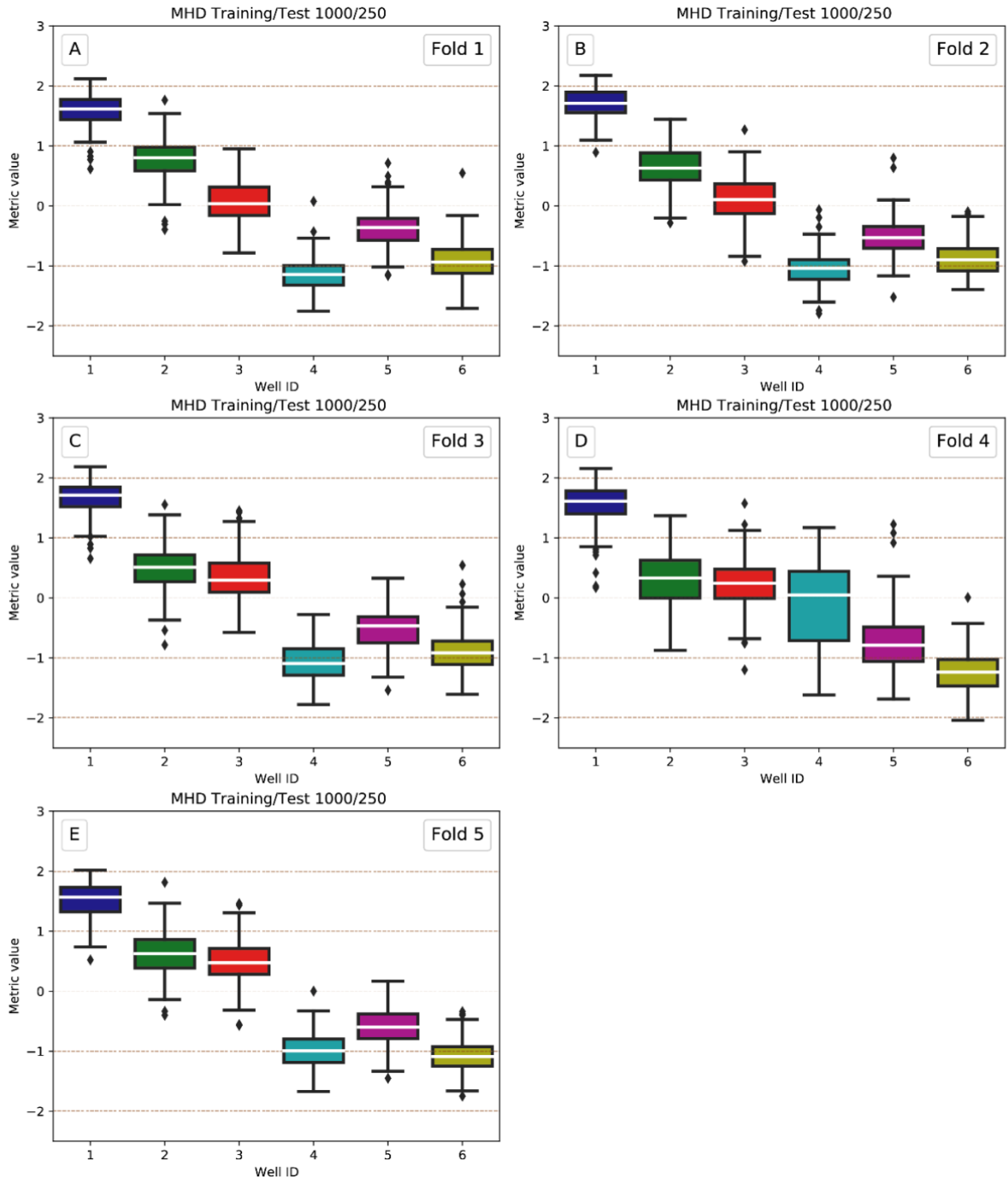

*Figure 12. Boxplots of the standardized MHD distance for each well and the 5 successive k-fold for a training dataset of size **1000** and **250** test samples. The different boxplots are consistent with each other across folds.*

## 5. Discussion

The main limitation of the BEL framework is to find an appropriate relationship between data and prediction. The algorithm used in this work (CCA) is based on linear combinations between predictor and target variables (in reduced-dimension space) as initially designed in BEL (Scheidt et al., 2018; Hermans et al., 2018). It was found to be sufficient in this work, but previous works have shown that highly non-linear relationships might still be difficult to characterize, which might introduce a bias in the prediction (Hermans et al., 2019), although CCA can model some non-linearity by using several dimensions. Recent works suggested that this can be handled by some adaptations of the learning procedure introducing some iterative update of the prior (Hermans et al., 2019; Michel et al. 2020b; Park and Caers, 2020). However, such adaptation increases the computation cost and reduces the adaptability of BEL has the iterative process is necessarily dependent on the data set, making it less efficient for experimental design.

Another limitation is the subjective choice of the size of the training set. Michel et al. (2020a) have shown that there is a threshold above which increasing the size does not affect the posterior prediction. This threshold is case-dependent. In this study, we found that a training set of 400 is sufficient for both prediction and experimental design purposes. For experimental design, it is shown through cross-validation that the size of the test set needed is at least 250. Additional variables for experimental design needed to be set by the practitioner are the number of PC's to keep for both predictor and target, the number of posterior samples to compute the uncertainty reduction, as well as the definition of the data-utility function, depending on the nature of the target variable.

A possible limitation of the approach is the heterogeneity of the medium used for forward modelling (e.g., hydraulic conductivity field). In this work, the different field are generated through sequential Gaussian simulations (2-points statistics) and are inherently smooth. The natural variability of geological media can be far from smooth, as for example a channelized medium (e.g., Lopez-Alvis et al., 2021), which requires more advanced simulation techniques such as multiple-point statistics (Mariethoz and Caers, 2015). However, BEL has been successfully used in such complex environments as well (Hermans et al., 2016; Yin and Caers, 2020), as what matters is the complexity of the target. Similarly, Hermans et al. (2018, 2019) have applied BEL with more uncertain components such as boundary conditions or spatial correlation parameters and 3D case with layers with different distributions. Such larger prior uncertainty would necessarily increase posterior uncertainty, but our experimental design approach could still be applied without loss of generality.

## 6. Conclusion

In this contribution, we propose a novel approach by combining experimental design with Bayesian Evidential Learning (BEL). The targets, wellhead protection areas (WHPA's) around a pumping well, are stochastically predicted by using as predictors breakthrough curves (BC's) from tracing experiments. Using a training set of small size (400 samples), a direct relationship between the predictor and the target is found and is used to estimate the full posterior distribution of an unknown target given any new predictor not included in the training set, corresponding to actual observed data. The prediction procedure comes at a relatively low computational costs and does not need any model calibration by data inversion.

We assess the informative content of the considered injection wells (data sources) in order to identify the optimal location of data sources by estimating the prior to posterior uncertainty reduction for series of prospective datasets. The uncertainty of the prediction is quantified by data-utility functions based on the Modified Hausdorff Distance (MHD) and the Structural Similarity (SSIM) index. We use a k-fold cross-validation procedure to validate the size of the prospective data sets. It is shown that a training set of 400 samples is sufficient both for estimating the WHPA with BEL and for experimental design purposes, combined with a test set of 250 samples required to make robust conclusions on the most informative data sources. Compared to previous approaches used for similar experimental design problems (e.g., Bayesian model averaging, surrogate modelling), our approach does not require several thousands (or more) forward model evaluations while we estimate the full posterior distribution of the target of interest without having to simplify the mathematical model.